\documentclass[10pt,twocolumn,letterpaper]{article}

\usepackage{cvpr}
\usepackage{times}
\usepackage{epsfig}
\usepackage{graphicx}
\usepackage{amsmath}
\usepackage{amssymb}
\usepackage{subfig}

\newcommand{\elu}{\operatorname{ELU}}
\newcommand{\elua}{\operatorname{CELU}}

\usepackage[breaklinks=true,bookmarks=false]{hyperref}

\cvprfinalcopy 

\def\cvprPaperID{****} 

\begin{document}

\title{Continuously Differentiable Exponential Linear Units}

\author{Jonathan T. Barron\\
{\tt\small barron@google.com}
}

\maketitle

\begin{abstract}
Exponential Linear Units (ELUs) are a useful rectifier for constructing deep learning architectures, as they may speed up and otherwise improve learning by virtue of not have vanishing gradients and by having mean activations near zero \cite{ELU}.
However, the ELU activation as parametrized in \cite{ELU} is not continuously differentiable with respect to its input when the shape parameter $\alpha$ is not equal to $1$.
We present an alternative parametrization which is $C^1$ continuous for all values of $\alpha$, making the rectifier easier to reason about and making $\alpha$ easier to tune.
This alternative parametrization has several other useful properties that the original parametrization of ELU does not: 1) its derivative with respect to $x$ is bounded, 2) it contains both the linear transfer function and ReLU as special cases, and 3) it is scale-similar with respect to $\alpha$.
\end{abstract}

\begin{figure*}
	\subfloat[][ $\elu(x, \alpha)$  \label{subfig:loss1}]{
		\includegraphics[width=3.3in]{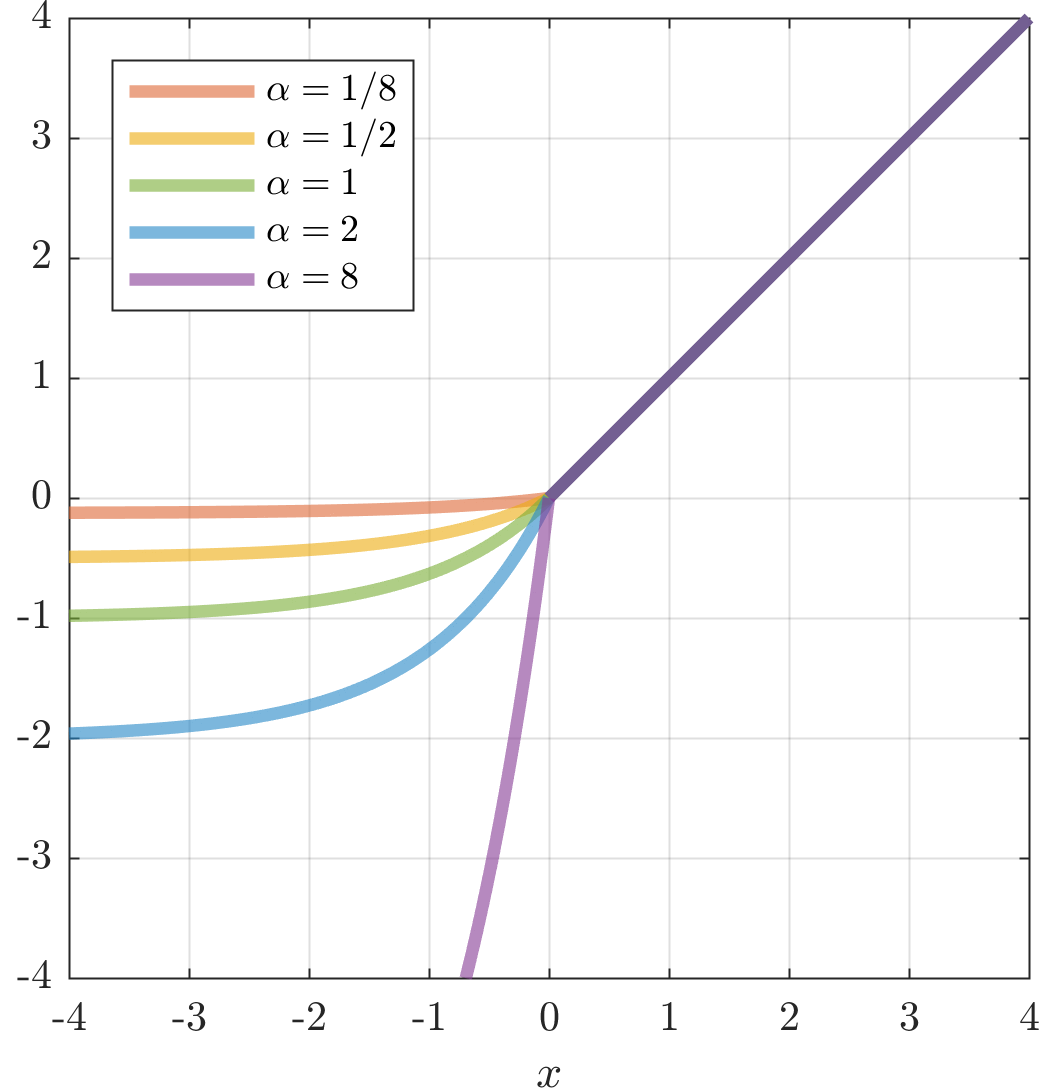}
	}
	\subfloat[][ ${d \over dx} \elu(x, \alpha)$  \label{subfig:grad1} ]{
		\includegraphics[width=3.3in]{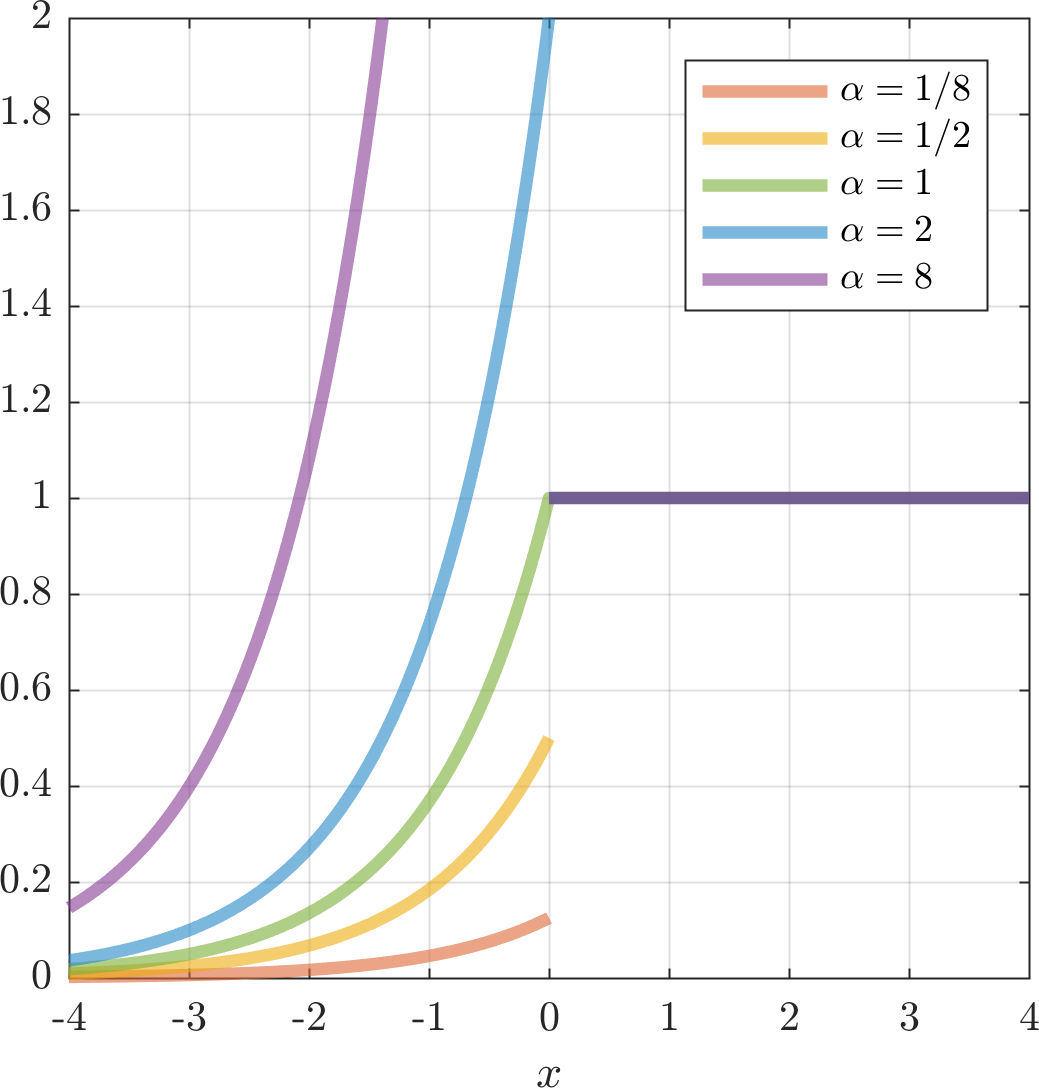}
	}
	\\
	\subfloat[][ $\elua(x, \alpha)$ ]{
		\includegraphics[width=3.3in]{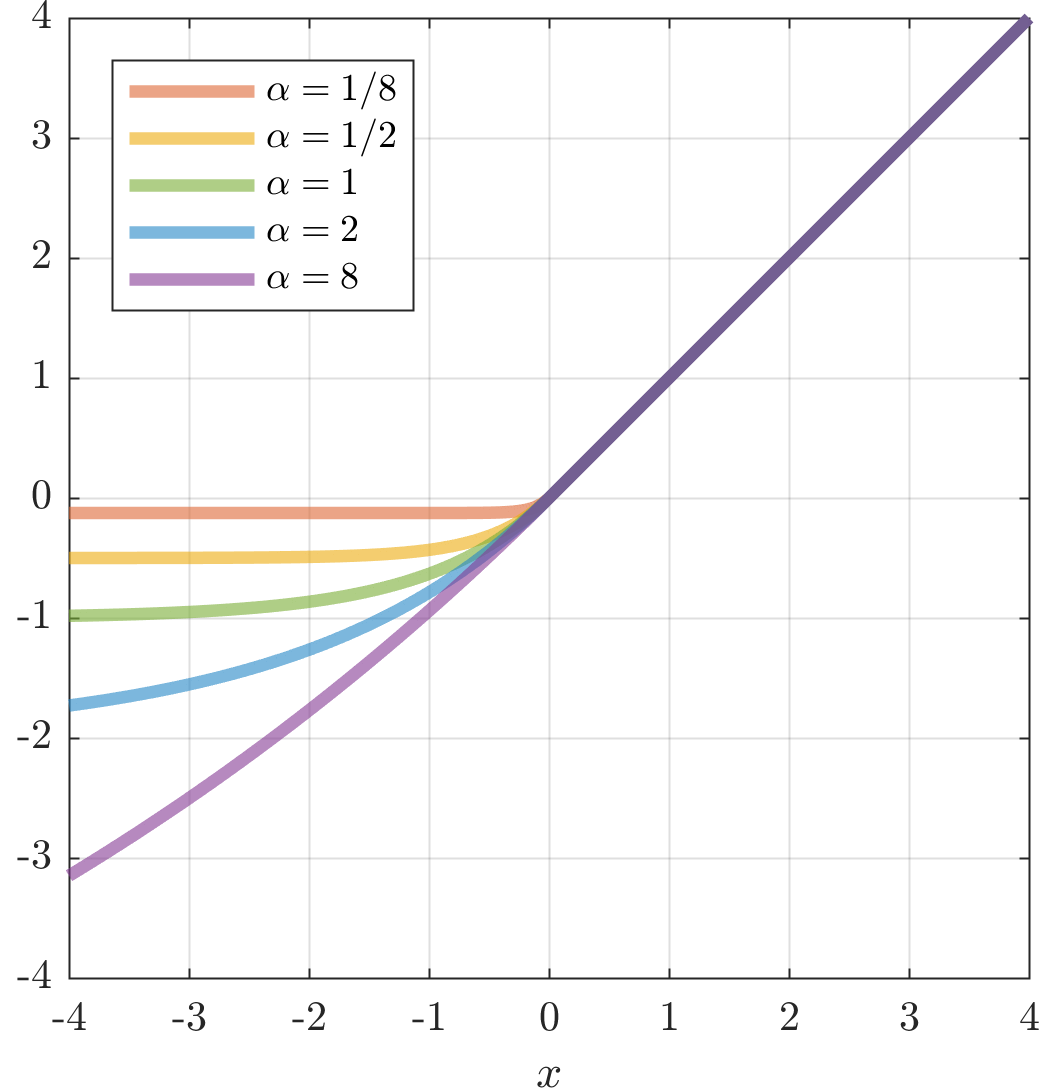}
	}
	\subfloat[][ ${d \over dx} \elua(x, \alpha)$ ]{
		\includegraphics[width=3.3in]{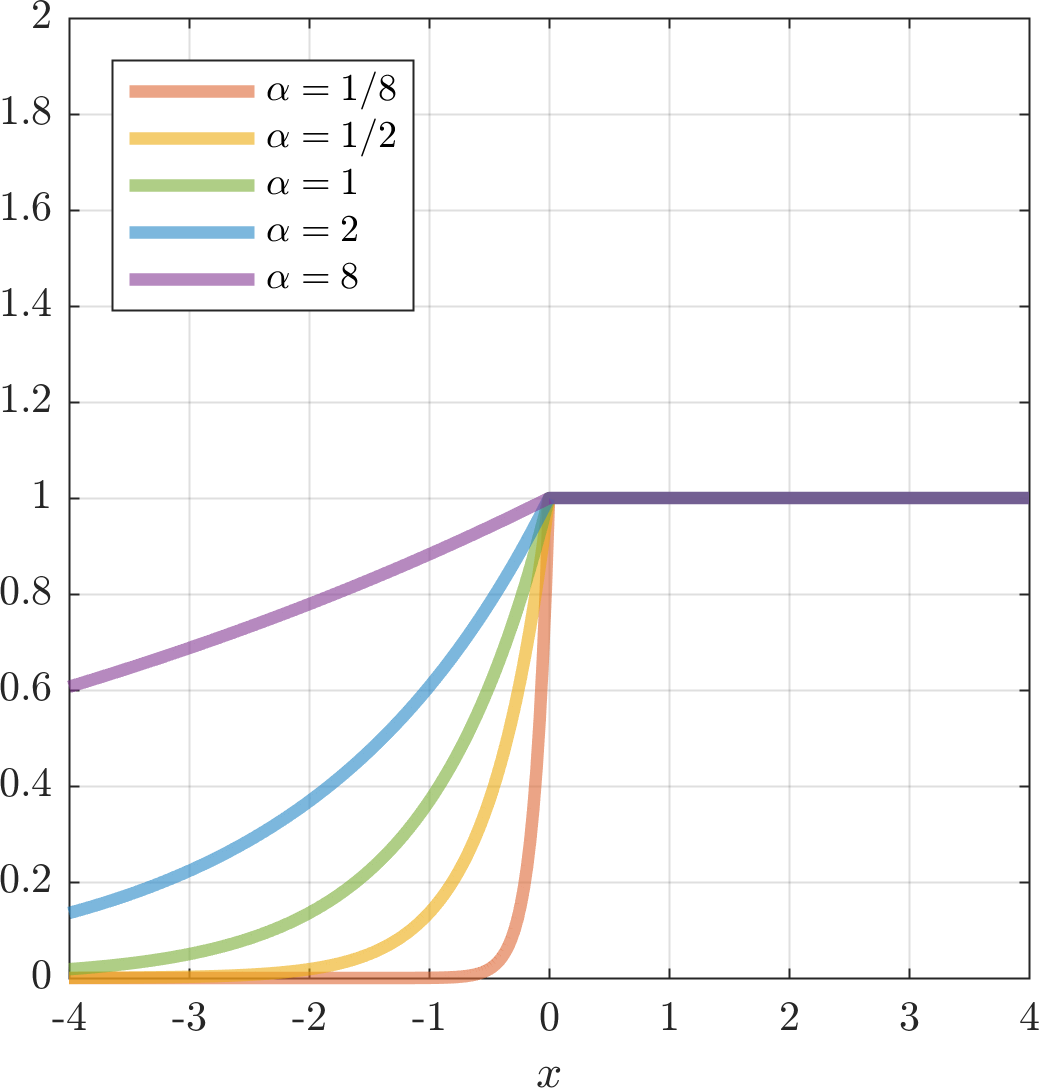}
	}
	\caption{
	The ELU activation function (top) as described in \cite{ELU} is not continuously differentiable with respect to $x$ for all value of $\alpha$.
	Our reparametrization (bottom) gives an activation function with the benefits of ELU, while being continuously differentiable, scale-similar, containing a linear function as a special case, and not having an unbounded derivative.}
\end{figure*}

The Exponential Linear Unit as described in \cite{ELU} is as follows:
\begin{align}
\elu(x, \alpha)={
	\begin{cases}
		x & {\mbox{if }}x\geq0 \\
		\alpha( \exp(x)-1) & {\mbox{otherwise}}
	\end{cases}}
\end{align}
Where $x$ is the input to the function, and $\alpha$ is a shape parameter.
The derivative of this function with respect to $x$ is:
\begin{align}
{d \over dx} \elu(x, \alpha)={
	\begin{cases}
		1 & {\mbox{if }}x\geq0 \\
		\alpha \exp(x) & {\mbox{otherwise}}
	\end{cases}}
\end{align}
In Figures~\ref{subfig:loss1} and \ref{subfig:grad1} we plot this activation and its derivative with respect to $x$ for different values of $\alpha$.
We see that when $\alpha \neq 1$, the activation's derivative is discontinuous at $x=0$.
Additionally we see that large values of $\alpha$ can cause a large (``exploding'') gradient for small negative values of $x$, which may make training difficult.

Our alternative parametrization of the ELU, which we dub ``CELU'', is simply the ELU where the activation for negative values has been modified to ensure that the derivative at $x=0$ for all values of $\alpha$ is $1$:
\begin{align}
\elua(x, \alpha)={
	\begin{cases}
		x & {\mbox{if }}x\geq0 \\
		\alpha \left( \exp \left( {x \over \alpha} \right) - 1 \right) & {\mbox{otherwise}}
	\end{cases}}
\end{align}
Note that ELU and CELU are identical when $\alpha = 1$:
\begin{equation}
 \forall_x \,\, \elu(x, 1) = \elua(x, 1)
\end{equation}
The derivative of the activation with respect to $x$ and $\alpha$ are as follows:
\begin{align}
{d \over dx} \elua(x, \alpha)&={
	\begin{cases}
		1 & {\mbox{if }} x \geq 0 \\
		\exp \left( {x \over \alpha} \right) & {\mbox{otherwise}}
	\end{cases}} \\
{d \over d\alpha} \elua(x, \alpha)&={
	\begin{cases}
		0 & {\mbox{if }} x \geq 0 \\
		\exp \left( {x \over \alpha} \right) \left(1 - {x \over \alpha} \right) - 1 & {\mbox{otherwise}}
	\end{cases}} \nonumber
\end{align}
Like in ELU, derivatives for CELU can be computed efficiently by precomputing $\exp \left( {x \over \alpha} \right)$ and using it for the activation and its derivatives.
Unlike ELU, CELU is scale-similar as a function of $x$ and $\alpha$:
\begin{equation}
\elua(x, \alpha) = {1 \over c} \elua(cx, c\alpha)
\end{equation}
The CELU also converges to $\operatorname{ReLU}$ as $\alpha$ approaches $0$ from the right and converges to a linear ``no-op'' activation as $\alpha$ approaches $\infty$:
\begin{align}
 \lim _{\alpha \to 0^{+}} \elua(x, \alpha) &= \max(0, x) \\
  \lim _{\alpha \to \infty} \elua(x, \alpha) &= x
\end{align}
This gives the CELU a nice interpretation as a way to interpolate between a ReLU and a linear function using $\alpha$.
Naturally, CELU can be slightly shifted in $x$ and $y$ such that it converges to any arbitrary shifted ReLU, in case negative activations are desirable even for small values of $\alpha$.

%
%
%

{\small
\bibliographystyle{ieee}
\bibliography{paper}
}

\end{document}